\documentclass{article}

\usepackage{microtype}
\usepackage{graphicx}
\usepackage{subcaption}
\usepackage{booktabs} 

\usepackage{hyperref}

 \usepackage[accepted]{aq_icml2026}

\usepackage{amsmath}
\usepackage{amssymb}
\usepackage{mathtools}
\usepackage{amsthm}
\usepackage{multirow}

\usepackage[capitalize,noabbrev]{cleveref}

\usepackage{graphicx}
\usepackage{tikz}
\usepackage{xcolor}
\definecolor{primary}{RGB}{41, 128, 185}
\definecolor{secondary}{RGB}{52, 152, 219}
\definecolor{accent}{RGB}{231, 76, 60}
\definecolor{success}{RGB}{46, 204, 113}
\definecolor{process}{RGB}{155, 89, 182}
\definecolor{light}{RGB}{236, 240, 241}
\usetikzlibrary{arrows.meta, positioning, shapes.geometric, shadows.blur, backgrounds, calc}

\theoremstyle{plain}

\theoremstyle{definition}

\theoremstyle{remark}

\usepackage[textsize=tiny]{todonotes}

\icmltitlerunning{When Uncertainty Isn't Enough: Self-Correction in Code Generation}

\begin{document}

\twocolumn[
  \icmltitle{When Uncertainty Isn't Enough: An Empirical Study of Self-Correction in Code Generation}



  \icmlsetsymbol{equal}{*}

  \begin{icmlauthorlist}
    \icmlauthor{Pranav Rakasi}{equal,yyy}
    \icmlauthor{Maanas Lalwani}{equal,xxx}
    \icmlauthor{Arnav Srivastava}{xxx}
    \icmlauthor{Arya Palanivel}{zzz}
    \icmlauthor{Tinuade Adeleke}{comp}
    \icmlauthor{Ruizhe Li}{bbb}
    \icmlauthor{Sean Wu}{aaa}
  \end{icmlauthorlist}

  \icmlaffiliation{yyy}{University of Michigan}
  \icmlaffiliation{xxx}{New York University}
  \icmlaffiliation{zzz}{University of Wisconsin-Madison}
  \icmlaffiliation{aaa}{University of Oxford}
  \icmlaffiliation{bbb}{University of Aberdeen}
  \icmlaffiliation{comp}{Algoverse AI}

  \icmlcorrespondingauthor{Tinuade Adeleke}{tinuade@algoverseairesearch.org}

  \icmlkeywords{Machine Learning, ICML}

  \vskip 0.3in
]



\printAffiliationsAndNotice{}  

\begin{abstract}
  Large language models for code generation often produce incorrect solutions without reliable indicators of failure. We study whether uncertainty estimation methods developed for natural language transfer to code generation, and whether such signals can improve code generation via selective self-correction. We evaluate five uncertainty methods: mean token entropy, verbalized confidence, $P(\text{True})$, entropy ensembles, and semantic entropy probes, across three small code LLMs on HumanEval and BigCodeBench. We find that multi-sample $P(\text{True})$ achieves the strongest correlation with correctness, while all the other methods, including semantic entropy probes, yield only weak correlation.
We then use these uncertainty signals to drive three self-correction policies: adaptive decoding, uncertainty-based regeneration, and verification-based regeneration. Our results reveal a stronger negative finding than anticipated: uncertainty-based self-correction fails to reliably improve Pass@1, degrading accuracy in 5 of 6 configurations across both benchmarks ($-3$pp to $-10$pp), and adaptive decoding degrades accuracy in 4 of 6 configurations. Only verification-based self-correction reliably improves Pass@1, with gains of $+6$ to $+26$ percentage points on HumanEval and $+8$ to $+20$ percentage points on BigCodeBench, scaling inversely with baseline strength. These findings replicate consistently across both benchmarks and suggest that cheap uncertainty estimators are insufficient on their own to improve code correctness, and that their practical value lies in serving as gating signals for costlier execution-based correction loops rather than as standalone substitutes for verification.
\end{abstract}

\section{Introduction}
Large Language Models (LLMs) have demonstrated impressive capabilities in code generation, often producing code that compiles and passes tests. However, their failures are unpredictable. A model that succeeds on one task may produce subtle bugs or completely incorrect solutions on another, without obvious warning signs to the user. Current code assistant systems typically treat every generation equally, regardless of how confident the model is in that solution. This means that a risky, low confidence code completion is delivered to the user just as readily as a high confidence one. The consequence is wasted computation and user frustration when faced with broken code.

A key insight is that incorporating uncertainty estimation into the code generation loop could enable an LLM-based coding assistant to be ``aware'' of its own reliability and self-correct before presenting the output to the user. This would mean the model evaluates how likely its generated code is to be correct. If confidence is high, the code can be emitted directly; if the model detects high uncertainty in its answer, the system can automatically revise or regenerate the code before the user ever sees a faulty solution.

Estimating the uncertainty of generated code has several potential benefits beyond coding assistant systems. In current practice, validating code correctness often relies on external checks like compiling and running the code against test cases. Although effective, such methods incur significant latency and are not always feasible. Uncertainty estimates offer a cheaper proxy for correctness, which is useful when we can tolerate some accuracy loss in exchange for speed or reduced resource use.

This tradeoff, sacrificing some accuracy for efficiency, is valuable in several settings. In reinforcement learning, where models are rewarded based on execution results, uncertainty estimation is much cheaper than execution-based rewards and could serve as a proxy for correctness. In interactive code completion assistants, real-time suggestions cannot wait for test execution; uncertainty could flag unreliable completions without running code at all. In batch refactoring or code review, uncertainty scores could prioritize which changes warrant human attention or testing.

This pattern appears in other domains as well. Medical diagnosis systems use uncertainty to escalate cases to specialists and autonomous vehicles use perception uncertainty to request human intervention. 

However, while uncertainty estimation has been studied in natural language tasks, it remains underexplored in code related tasks. Code has unique semantic properties; a single wrong token can cause complete failure, two nearly identical snippets may behave entirely differently, while two syntactically distinct programs may be functionally equivalent. These properties suggest that uncertainty techniques from natural language may not transfer directly, motivating code specific investigation. 

In this work, we address two questions. First, do uncertainty estimation techniques developed for natural language transfer effectively to code generation? Second, can uncertainty estimates improve the performance of LLM-based coding assistants while adding minimal latency, serving as a cheap proxy for correctness?

Our contributions are as follows:
\begin{itemize}
    \item We provide the first systematic comparison of five uncertainty estimation methods for code generation, showing that single-forward-pass estimators, including semantic entropy probes, yield only weak correlation with correctness, while verbalized confidence is unreliable for small models. Multi-sample $P(\text{True})$ is the only method to achieve strong correlation, but at substantially higher cost.
    \item To our knowledge, we provide one of the first systematic comparisons of five uncertainty estimation methods for code generation, evaluated on \textbf{both HumanEval and BigCodeBench}. We show that single forward pass estimators, including semantic entropy probes, yield only weak correlation with correctness, while verbalized confidence is unreliable for small models. Multi-sample $P(\text{True})$ is the only method to achieve strong correlation, but at substantially higher cost

    \item We propose an uncertainty-driven self-correction framework with three correction policies and two generation strategies: full function regeneration (SLT) and proactive regeneration (TBG), the latter using pre-generation hidden states to gate the decoding strategy before any tokens are committed.
    \item We report a strong negative result: uncertainty-based self-correction degrades Pass@1 in 5 of 6 configurations tested across HumanEval and BigCodeBench, and adaptive decoding degrades accuracy in 4 of 6 configurations. This challenges the intuition that uncertainty-aware decoding and uncertainty-guided regeneration should help, and suggests that weak uncertainty signals are not merely uninformative but actively harmful when used as standalone correction triggers.
    \item We demonstrate that verification-based self-correction is the only reliably beneficial policy, suggesting that the practical value of uncertainty lies not in replacing execution feedback but in selectively triggering it.
\end{itemize}

\begin{figure}[t]
\centering
\resizebox{\columnwidth}{!}{%
\begin{tikzpicture}[
    font=\sffamily\small,
    node distance=5mm and 8mm,
    block/.style={
        rectangle, draw=primary!80, fill=primary!10,
        rounded corners=3pt, align=center, 
        minimum width=3.8cm, minimum height=1.1cm, line width=0.8pt,
        blur shadow={shadow blur steps=5, shadow xshift=0pt, shadow yshift=-2pt, shadow blur radius=3pt, shadow opacity=25}
    },
    wideblock/.style={
        rectangle, draw=secondary!80, fill=secondary!15,
        rounded corners=3pt, align=center, 
        minimum width=4.5cm, minimum height=1.2cm, line width=0.8pt,
        blur shadow={shadow blur steps=5, shadow xshift=0pt, shadow yshift=-2pt, shadow blur radius=3pt, shadow opacity=25}
    },
    processblock/.style={
        rectangle, draw=process!80, fill=process!12,
        rounded corners=3pt, align=center, 
        minimum width=4.5cm, minimum height=1.2cm, line width=0.8pt,
        blur shadow={shadow blur steps=5, shadow xshift=0pt, shadow yshift=-2pt, shadow blur radius=3pt, shadow opacity=25}
    },
    decision/.style={
        diamond, draw=accent!80, fill=accent!12,
        align=center, inner sep=2pt, aspect=2.2, line width=0.8pt,
        blur shadow={shadow blur steps=5, shadow xshift=0pt, shadow yshift=-2pt, shadow blur radius=3pt, shadow opacity=25}
    },
    acceptblock/.style={
        rectangle, draw=success!80, fill=success!12,
        rounded corners=3pt, align=center, 
        minimum width=3.5cm, minimum height=1.0cm, line width=0.8pt,
        blur shadow={shadow blur steps=5, shadow xshift=0pt, shadow yshift=-2pt, shadow blur radius=3pt, shadow opacity=25}
    },
    finalblock/.style={
        rectangle, draw=primary!90, fill=light,
        rounded corners=3pt, align=center, 
        minimum width=4.5cm, minimum height=1.0cm, line width=1.2pt,
        blur shadow={shadow blur steps=5, shadow xshift=0pt, shadow yshift=-2pt, shadow blur radius=4pt, shadow opacity=30}
    },
    line/.style={-{Latex[length=2.5mm, width=2mm]}, line width=1pt, color=primary!70},
    decisionline/.style={-{Latex[length=2.5mm, width=2mm]}, line width=1pt, color=accent!70},
    labelstyle/.style={font=\sffamily\footnotesize\bfseries, fill=white, inner sep=2pt, rounded corners=1pt}
]

\node[block] (prompt) {\textbf{Problem Prompt}\\{\footnotesize HumanEval / BigCodeBench}};
\node[wideblock, below=of prompt] (llm) {\textbf{Base LLM} \textit{(single run)}\\{\footnotesize Generates code + internals}};
\node[block, below left=of llm] (code) {\textbf{Code Output}};
\node[block, below right=of llm] (states) {\textbf{Logits / Hidden States}};
\node[processblock, below=of states, xshift=-2.35cm] (unc) {\textbf{Uncertainty Estimation}\\[-2pt]
{\footnotesize Semantic Entropy Probe}\\[-1pt]
{\scriptsize (TBG / SLT)}};
\node[decision, below=of unc] (decide) {\small $u(x) > \tau$?};
\node[acceptblock, below left=of decide, xshift=-6mm] (accept) {\textbf{Return Output}\\{\footnotesize (no extra cost)}};
\node[processblock, below right=of decide, xshift=6mm] (policy) {\textbf{Self-Correction Policy}\\[-2pt]
{\footnotesize (2a) Regenerate}\\[-1pt]
{\footnotesize (2b) Adaptive decode}};
\node[finalblock, below=of decide, yshift=-20mm] (final) {\textbf{Final Code Output}};

\begin{scope}[on background layer]
    \fill[light!30, rounded corners=5pt] ($(prompt.north west)+(-0.3,0.3)$) rectangle ($(final.south east)+(0.3,-0.3)$);
\end{scope}

\draw[line] (prompt) -- (llm);
\draw[line] (llm) -- (code);
\draw[line] (llm) -- (states);
\draw[line] (states) -- (unc);
\draw[line] (unc) -- node[labelstyle, right, xshift=1mm] {$u(x)$} (decide);
\draw[decisionline] (decide) -- node[labelstyle, above] {No} (accept);
\draw[decisionline] (decide) -- node[labelstyle, above] {Yes} (policy);
\draw[line] (accept) |- (final);
\draw[line] (policy) |- (final);

\end{tikzpicture}
}
\caption{Overview of our uncertainty-aware code generation framework. The model produces an initial solution and intermediate representations. Uncertainty estimators compute a scalar score $u(x)$, which triggers a self-correction policy only when uncertainty exceeds a threshold $\tau$.}
\label{fig:overview}
\end{figure}
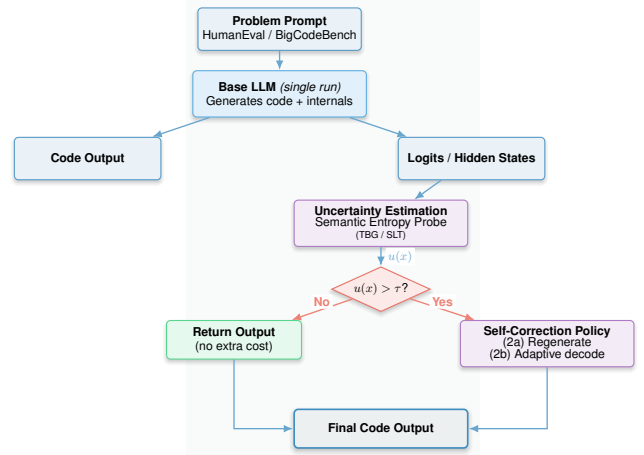

\section{Background and Related Works}

\textbf{Uncertainty Estimation in LLMs}: For comprehensive surveys on uncertainty estimation in LLMs, we refer to \cite{tao2025revisitinguncertaintyestimationcalibration, Huang_2025, xia2025surveyuncertaintyestimationmethods}. We briefly discuss three categories explored in this work.

The first leverages internal probability signals from decoding. Mean Token Entropy (MTE) computes Shannon entropy over the model's output distribution at each generation step, aggregated across the sequence \cite{fomicheva2020unsupervised}, and has become a common baseline for hallucination detection and reliability scoring \cite{Huang_2025, fadeeva-etal-2023-lm, vashurin-etal-2025-benchmarking}. However, \cite{kuhn2023semanticuncertaintylinguisticinvariances} observed that token-level probabilities estimate uncertainty poorly since different token sequences may convey the same meaning, and proposed Semantic Entropy, which clusters multiple generated answers by meaning and computes entropy over these clusters. \cite{li2025semanticvolumequantifyingdetecting} similarly uses the Gram matrix determinant of semantic embeddings to quantify dispersion. Both methods improve uncertainty estimation but require multiple generations. \cite{kossen2024semanticentropyprobesrobust} addresses this by training linear probes to approximate semantic entropy from hidden states of a single generation, reducing overhead to nearly zero with comparable performance.

In the second category, the model verbally expresses confidence. Verbalized Confidence Scores instruct LLMs to report confidence alongside answers \cite{tian2023justask, lin2022teaching, xiong2023express, kadavath2022language}. This requires no access to hidden states, but reliability is contested: some studies report reasonable calibration \cite{tian2023justask, lin2022teaching}, while others find systematic overconfidence \cite{xiong2023express, kadavath2022language}. \cite{yang2024verbalized} provides the most systematic investigation, showing that prompt design dominates calibration quality and that small LLMs produce nearly uninformative scores, while larger models benefit from well designed prompts. The self-evaluation pattern of asking for probability of correctness is commonly known as P(True).

The third category uses ensembling. \citet{tonolini2024bayespe} introduces Bayesian Prompt Ensembles, computing output probabilities through weighted ensembles of semantically equivalent prompts via Bayesian variational inference. While achieving superior calibration in zero and few-shot classification, it requires multiple LLM calls per input and is difficult to apply to long-form generation like code. In this work, we adapt this method with design choices to control cost and define task-appropriate uncertainty targets.

\textbf{Uncertainty Estimation for Code Generation:} \cite{sharma2025assessingcorrectnessllmbasedcode} adapted entropy and mutual-information methods for code tasks, finding a weak negative correlation between uncertainty scores and correctness but showing that uncertainty-based abstention can reduce incorrect outputs to near-zero. We build on this direction but use uncertainty as a self-correction rather than abstention policy. \cite{zhu2025uncertaintyguidedchainofthoughtcodegeneration} propose UnCert-CoT, which monitors entropy at line boundaries and switches to chain-of-thought decoding when uncertainty is high ($+6.1\%$ on MHPP). \cite{he2025bettercodegenerationadaptive} introduce AdaDec, which pauses at high-entropy tokens for lookahead reranking ($~15.5\%$ gains on HumanEval/MBPP). While both leverage entropy for token- or line-level decisions, our work studies uncertainty at the semantic level of entire generations, using estimates as control signals for higher-level self-correction policies with minimal additional latency.

\section{Methodology}

Our methodology consists of two components: 1.) A systematic evaluation of uncertainty estimation methods for code generation, and 2.) An uncertainty driven self-correction framework that uses these signals to selectively trigger regeneration.

\subsection{Problem Setup}
Let $x$ denote a programming problem (e.g., a HumanEval prompt), and let $y \sim p_\theta(y \mid x)$ be a code solution generated by a language model $\theta$. 
Each generated solution is evaluated by a functional correctness oracle (unit tests), yielding a binary outcome $c(y) \in \{0,1\}$. Our goal is to compute an uncertainty score $U(x)$ for each generated solution such that higher uncertainty correlates with a higher probability of failure, and to use this signal to improve code generation via selective self-correction.

\subsection{Uncertainty Estimation Evaluation}
We evaluate multiple uncertainty estimation techniques discussed in detail below on a code generation dataset. All methods output a scalar uncertainty score per problem instance. We evaluate correctness of the generated code using the verifiers library \cite{brown_verifiers_2025}.

\subsubsection{Mean Token Entropy}
\label{sec:mte}
Mean Token Entropy (MTE) quantifies the average uncertainty in a model's token predictions during generation. For each generated token, entropy is computed from the model's output probability distribution over the vocabulary: 

\[
H = -\sum_{i} p_i \log(p_i),
\]

where $p_i$ is the probability of token $i$. In practice, we extract logits from the model at each generation step, apply softmax to obtain probabilities, and compute entropy. The Mean Token Entropy is the arithmetic mean of these per-token entropies across all generated tokens. MTE is reported in nats (natural units, base $e$). Lower MTE indicates lower uncertainty (more peaked distributions), while higher MTE indicates higher uncertainty (more uniform distributions).

\subsubsection{Verbalized Confidence Scores}

 In this approach the model is explicitly prompted to output a numeric confidence score alongside its code. Concretely, for each HumanEval prompt $X$, the model produces a completion $Y$ that contains (i) a Python solution and (ii) a confidence statement of the form \texttt{Confidence: c}, where $c \in [0,1]$ is intended to represent the model's self-assessed probability that the solution will pass all tests. We extract this value with pattern-based parsing and clamp it to the valid range:

\[
\hat{c} = \min\bigl(1,\max(0,c)\bigr).
\]

When multiple confidence-like patterns occur, we use the last successfully parsed value. If no confidence can be parsed, we treat $\hat{c}$ as missing and the stopping condition cannot be satisfied for that attempt.



\subsubsection{P(True)}

We estimate the uncertainty of the model using the probability of correctness, denoted as \emph{P(True)}, grounded in external verification. Let $p_\theta(c \mid x)$ denote a language model with parameters $\theta$ generating a candidate solution $c$ for a task prompt $x$. For each task, we sample $K$ independent candidate solutions using stochastic decoding:

\[
\{\hat{c}_1, \ldots, \hat{c}_K\} \sim p_\theta(c \mid x).
\]

Each candidate is evaluated using an external verifier $V(\cdot)$, which returns a binary correctness signal:

\[
V(\hat{c}_k) =
\begin{cases}
1 & \text{if the solution passes all tests}, \\
0 & \text{otherwise}.
\end{cases}
\]

We define verifier-based \emph{P(True)} as the empirical success probability:

\[
P(\text{True} \mid x)
=
\frac{1}{K}
\sum_{k=1}^{K}
V(\hat{c}_k).
\]

This quantity estimates the probability that the model produces a correct solution under stochastic sampling. Unlike self-evaluated confidence, which reflects linguistic belief, verifier-based \emph{P(True)} measures observable task correctness and is therefore better calibrated.


\subsubsection{Entropy Ensembles via Prompt Diversity}

For each HumanEval task, we define a fixed set of $N$ semantically equivalent system prompts that differ in phrasing and emphasis but preserve task intent. For a given input $x$, the model generates exactly one completion per prompt using stochastic decoding (temperature $= 0.9$); variation across the $N$ completions therefore reflects both prompt-induced and sampling-induced uncertainty. For each prompt-conditioned completion we compute Mean Token Entropy (MTE) as described in Section~\ref{sec:mte}, yielding a set of uncertainty values $\{\mathrm{MTE}_i(x)\}_{i=1}^{N}$. We summarize prompt-induced uncertainty using two statistics:
\[
H(x) = \frac{1}{N}\sum_{i=1}^{N} \mathrm{MTE}_i(x)
\]
\[
V(x) = \frac{1}{N}\sum_{i=1}^{N}\big(\mathrm{MTE}_i(x) - H(x)\big)^2.
\]
Here, $H(x)$ captures the model's average uncertainty across prompts, while $V(x)$ measures disagreement in uncertainty across prompts and serves as a proxy for epistemic instability. Intuitively, low values of both $H(x)$ and $V(x)$ indicate consistent, confident behavior across prompts, whereas large variance suggests prompt-sensitive reasoning. To calibrate these statistics against functional correctness, we train a lightweight logistic regression model on a validation set to predict failure probability:
\[
U(x) = P_{\mathrm{fail}}(x) = \sigma\!\left(w_1 H(x) + w_2 V(x) + b\right),
\]
where $\sigma(\cdot)$ denotes the sigmoid function. The resulting scalar $U(x)$ serves as the uncertainty estimate.

\subsubsection{Semantic Entropy Probes (SEPs)}

We implement Semantic Entropy Probes (SEPs) to predict uncertainty from internal model representations. For each prompt $x$, we generate multiple solutions and cluster them by functional behavior using symbolic execution \cite{Ball2015DeconstructingDS}. Let the resulting semantic clusters have probabilities $\{p_j\}$, computed by aggregating token-level log-probabilities with length normalization. 
Semantic entropy is then:
\[
H_{\text{sem}}(x) = -\sum_j p_j \log p_j.
\]

We train multiple classifiers including a linear logistic regression model and a small neural network using model hidden states as input and a high/low semantic entropy label as the target. We consider two representations:
\begin{itemize}
    \item \textbf{SLT (Second-to-Last Token):} captures uncertainty after solution generation.
    \item \textbf{TBG (Token-Before-Generation):} captures uncertainty before any tokens are generated.
\end{itemize}

At test time, a single hidden state is extracted and passed through the probe to obtain an uncertainty score.

\subsection{Uncertainty Driven Self Correction}
\label{self-correction}

\subsubsection{Full Function Regeneration}
Given an initial solution with uncertainty $U(x)$ from the SLT probe, we trigger regeneration when $U(x) > \tau$. We evaluate two correction policies: (1)~\textit{uncertainty-based self-correction}, which regenerates using stochastic decoding until $U(x) \leq \tau$, and (2)~\textit{verification-based self-correction}, which regenerates until the solution passes all test cases. Both policies are capped at $N{=}5$ attempts; if neither stopping condition is met, the attempt with the lowest uncertainty (policy~1) or a passing result (policy~2) is returned. Thresholds are selected per-model via grid search over a HumanEval validation split, optimizing F1 to balance unnecessary corrections against missed failures (details in Appendix~\ref{app:probe-design}).

\subsubsection{Proactive Regeneration}
Rather than generating a full solution before assessing uncertainty, we use the TBG probe to estimate uncertainty from the model's hidden state after producing only a single next token. If uncertainty exceeds $\tau$, the model switches to adaptive decoding for the remainder of generation; otherwise, it proceeds with standard greedy decoding. This combines token-level and semantic-level entropy signals in a single lightweight decision, avoiding the latency cost of full regeneration when confidence is high.

\subsection{Evaluation}
We evaluate on HumanEval and BigCodeBench. Models include LLaMA 3, Qwen2.5-Coder, and DeepSeek-R1 ($\sim$3B instruct variants). Uncertainty quality is measured via correlation with pass/fail outcomes (Pearson correlation) and calibration plots. Self-correction is evaluated using pass@k and wall-clock latency.

\section{Results and Discussion}

\subsection{Uncertainty Evaluations}
\label{sec:uncertainty_evaluations}
The correlation analysis in Table~\ref{tab:uncertainty_correlations} reveals that P(True) achieves the strongest performance across all three models ($0.842$, $0.782$, $0.303$ for Llama, Qwen, and DeepSeek respectively), though the substantial degradation on DeepSeek suggests that smaller models exhibit less consistent self-evaluation behavior. All the other methods yield weak negative correlations with correctness, in the range of $-0.111$ to $-0.631$. Semantic entropy probes, in particular, place in the same regime as the simpler entropy-based baselines ($-0.196$ to $-0.475$). 

These results carry a clear practical implication. Multi-sample P(True) remains the strongest available uncertainty signal but requires $K$ stochastic rollouts per task, making it expensive for real-time use. Single forward pass alternatives are an order of magnitude cheaper but, on these benchmarks, deliver substantially weaker correlation with correctness. We nonetheless adopt semantic entropy probes
for our self-correction experiments because they offer the lowest overhead signal and allow us to isolate the question of whether \emph{any} cheap uncertainty estimator, even an imperfect one, can drive useful corrective behavior. The results in Section~\ref{sec:self_correction} should therefore be read as a test of whether weak but cheap uncertainty signals are actionable, not as a test of the
strongest possible uncertainty estimator.

\begin{table*}[t]
\centering
\small

\begin{tabular}{lccc}
\toprule
\textbf{Method} & \textbf{Llama-3.2-3B-Instruct} & \textbf{Qwen2.5-Coder-3B-Instruct} & \textbf{Deepseek-Coder-1.3B-Instruct} \\
\midrule
MTE & -0.416 & -0.234 & -0.216 \\
Verbalized Confidence & -0.327 & -0.090 & -0.114 \\
P(True) & \textbf{0.842} & 0.782 & 0.303 \\
Entropy Ensemble & -0.631 & -0.476 & 0.021 \\
Semantic Entropy Probe & -0.196 & -0.475 & -0.416 \\
\bottomrule
\end{tabular}
\caption{Correlation Analysis of Uncertainty Estimation Methods on the HumanEval dataset}
\label{tab:uncertainty_correlations}
\end{table*}

\subsection{Self-Correction Using Uncertainty}
\label{sec:self_correction}

We evaluate the accuracy of our semantic entropy probes (SEP) before applying them in uncertainty-guided decoding; Appendix~\ref{app:probe-design} reports probe classification performance and details the threshold tuning experiments. Table~\ref{tab:main_results} presents Pass@1 accuracy and wall-clock latency for each self-correction method across three code generation models on HumanEval. We compare three uncertainty-guided policies: adaptive decoding, uncertainty-based self-correction, and verification-based self-correction, across two generation strategies: full-function regeneration (SLT), which assesses uncertainty over a complete output, and proactive regeneration (TBG), which uses the hidden state after a single next-token prediction to gate the decoding strategy before committing to the rest of the sequence.

Adaptive decoding degraded Pass@1 in five of six HumanEval configurations, with degradations ranging from $-3.23$pp to $-6.45$pp. The single positive result, Llama-3.2-3B under TBG gating ($+3.23$pp), suggests that pre-generation uncertainty gating can occasionally help when the base model is weak and the gating decision is made once rather than repeatedly throughout decoding. The dominant pattern, however, remains degradation. Our adaptive decoding experiments use learned entropy thresholds calibrated on the BigCode dataset, following the approach of \citet{he2025bettercodegenerationadaptive}. Despite their claim of generalizability, we observe broad degradation when transferring these thresholds to HumanEval, suggesting that threshold calibration is sensitive to distributional differences between training and evaluation datasets and underscoring the need for more robust, dataset-agnostic threshold selection.

Uncertainty-based self-correction degraded Pass@1 in every HumanEval configuration tested ($-3.23$pp to $-9.68$pp). This is a strong negative result and indicates that resampling guided purely by probe-estimated uncertainty tends to replace correct greedy outputs with incorrect stochastic ones. The probe signal, while cheap, is not selective enough to identify which high-uncertainty greedy outputs would actually benefit from regeneration; in aggregate, the cost of corrupting correct generations outweighs the benefit of fixing incorrect ones. This finding is consistent with the weak correlations reported in Section~\ref{sec:uncertainty_evaluations}: a signal that correlates only weakly with correctness cannot reliably distinguish recoverable failures from successful generations that merely look uncertain.

Verification-based self-correction, available only under the full-function regeneration pipeline, was the only method to improve Pass@1 in every HumanEval configuration, though the magnitude of improvement varied substantially across models. DeepSeek-Coder-1.3B saw the largest absolute gain ($+25.81$ percentage points, from $0.61$ to $0.87$ at $2.4\times$ latency overhead), Llama-3.2-3B improved moderately ($+6.45$ percentage points, from $0.38$ to $0.45$ at $4.4\times$), and Qwen2.5-Coder-3B improved by approximately $6$ percentage points (from $0.81$ to $0.87$ at $1.9\times$). This pattern suggests that verification-based correction yields the largest gains where the base model is weakest and has the most headroom, while ceiling effects limit gains for stronger models that already produce correct greedy outputs on most tasks.

Proactive regeneration (TBG) was originally motivated as a low-latency alternative to full-function regeneration, on the intuition that gating the decoding strategy from a single pre-generation hidden state would avoid the cost of generating and regenerating complete solutions. In our measurements, however, TBG adaptive decoding incurs $2.8\times$ to $10.1\times$ baseline latency; in two of three models, this exceeds the cost of SLT verification self-correction. The latency advantage of proactive gating is therefore not realized in our current implementation, likely because the adaptive policy itself is expensive once triggered.

Our initial hypothesis that uncertainty estimation could serve as a cheap proxy for correctness, reducing dependence on execution-based verification, is not supported by these results. While probe-based uncertainty shows weak correlation with correctness (Section~\ref{sec:uncertainty_evaluations}), using these signals alone to guide correction consistently degrades Pass@1. Verification-based self-correction, which incorporates test execution feedback, is the only policy that reliably improves accuracy across all models. This points to a re-framing of the role of uncertainty in code generation pipelines: uncertainty signals and execution verification are complementary rather than substitutable, and the practical value of cheap uncertainty estimators likely lies not in replacing execution feedback but in serving as a lightweight gating mechanism that decides when to invoke costlier verification-based correction. More broadly, these results highlight a distinction often elided in the uncertainty literature: even when a signal correlates with correctness, acting on it requires a corrective mechanism that is itself reliable. Without such a mechanism---and execution feedback is, in our experiments, the only reliable one---acting on uncertainty introduces harmful variability rather than reducing it.
\begin{table*}[!tbp]
\centering
\small
\setlength{\tabcolsep}{4pt}
\renewcommand{\arraystretch}{1.08}
\resizebox{\textwidth}{!}{%
\begin{tabular}{@{}lllcccc@{}}
\toprule
\textbf{Model} & \textbf{Feature} & \textbf{Method} & \textbf{Pass@1} & \textbf{Avg. Latency} & $\boldsymbol{\Delta}$\textbf{Acc} & \textbf{Latency $\times$} \\
\midrule
\multirow{3}{*}{Llama-3.2-3B} 
& \multirow{3}{*}{SLT}
& Baseline (greedy decoding) & 0.39& 1.81s & -- & 1.0$\times$ \\
& & Uncertainty Self-Correction & 0.29 & 8.41s & -9.68pp & 4.6$\times$ \\
& & Verification Self-Correction & 0.45 & 7.90s & +6.45pp & 4.4$\times$ \\
\midrule
Llama-3.2-3B & TBG & Adaptive Decoding & 0.42 & 18.22s & +3.23pp & 10.1$\times$ \\
\midrule
\multirow{3}{*}{Qwen2.5-Coder-3B}
& \multirow{3}{*}{SLT}
& Baseline (greedy decoding) & 0.81 & 4.09s & -- & 1.0$\times$ \\
& & Uncertainty Self-Correction & 0.77 & 16.85s & -3.23pp & 4.1$\times$ \\
& & Verification Self-Correction & 0.87 & 7.73s & +6.45pp & 1.9$\times$ \\
\midrule
Qwen2.5-Coder-3B & TBG & Adaptive Decoding & 0.77 & 15.73s & -3.23pp & 3.8$\times$ \\
\midrule
\multirow{3}{*}{DeepSeek-Coder-1.3B}
& \multirow{3}{*}{SLT}
& Baseline (greedy decoding) & 0.61 & 4.07s & -- & 1.0$\times$ \\
& & Uncertainty Self-Correction & 0.58 & 18.87s & -3.23pp & 4.6$\times$ \\
& & Verification Self-Correction & 0.87 & 9.80s & +25.81pp & 2.4$\times$ \\
\midrule
DeepSeek-Coder-1.3B & TBG & Adaptive Decoding & 0.55 & 11.48s & -6.45pp & 2.8$\times$ \\
\bottomrule
\end{tabular}%
}
\caption{Self-correction results on \textbf{HumanEval} ($n{=}31$ held-out evaluation problems. Adaptive decoding degrades Pass@1 in two of three configurations and uncertainty-based self-correction degrades it in all three, whereas verification-based self-correction is the only policy to improve Pass@1 in every configuration. ``Latency'' is mean wall-clock time per problem and ``Latency $\times$'' is relative to the greedy baseline.}
\label{tab:main_results}
\end{table*}

\paragraph{Generalization to BigCodeBench.}
To assess whether our findings generalize beyond HumanEval, we evaluate the same pipeline on BigCodeBench~\cite{bigcodebench}, a harder benchmark of 155 tasks drawn from real-world library usage (v0.1.4 split). Table~\ref{tab:bigcodebench_results} reports the results. Baseline Pass@1 values are substantially lower (0.18--0.30 vs.\ 0.38--0.81 on HumanEval), reflecting the greater difficulty of BigCodeBench tasks. Despite this distributional shift, the key findings replicate: uncertainty-based self-correction is unreliable ($-3.87$pp to $+5.81$pp, with the single positive result for the weakest baseline model), adaptive decoding mostly degrades performance ($-4.52$pp to $+0.65$pp), and verification-based self-correction is the only policy to reliably improve Pass@1 in every configuration, with gains of $+7.74$ to $+20.00$ percentage points. Notably, DeepSeek-Coder-1.3B again shows the largest absolute gain ($+20.00$pp), consistent with the pattern observed on HumanEval where weaker base models have more headroom for improvement. The single anomaly---uncertainty self-correction helping DeepSeek on BigCodeBench---likely reflects the very low baseline accuracy ($0.18$), which leaves substantial room for stochastic resampling to recover correct solutions even with a weak triggering signal. These results strengthen our main conclusion: verification feedback is the critical component for reliable self-correction, and this finding is robust across benchmarks of varying difficulty.

\begin{table*}[!tbp]
\centering
\small
\setlength{\tabcolsep}{4pt}
\renewcommand{\arraystretch}{1.08}
\resizebox{\textwidth}{!}{%
\begin{tabular}{@{}lllcccc@{}}
\toprule
\textbf{Model} & \textbf{Feature} & \textbf{Method} & \textbf{Pass@1} & \textbf{Avg. Latency} & $\boldsymbol{\Delta}$\textbf{Acc} & \textbf{Latency $\times$} \\
\midrule
\multirow{3}{*}{Llama-3.2-3B} 
& \multirow{3}{*}{SLT}
& Baseline (greedy decoding) & 0.271 & 5.57s & -- & 1.0$\times$ \\
& & Uncertainty Self-Correction & 0.239 & 16.97s & -3.23pp & 3.0$\times$ \\
& & Verification Self-Correction & 0.368 & 32.91s & +9.68pp & 5.9$\times$ \\
\midrule
Llama-3.2-3B & TBG & Adaptive Decoding & 0.252 & 20.51s & -2.58pp & 3.4$\times$ \\
\midrule
\multirow{3}{*}{Qwen2.5-Coder-3B}
& \multirow{3}{*}{SLT}
& Baseline (greedy decoding) & 0.297 & 15.99s & -- & 1.0$\times$ \\
& & Uncertainty Self-Correction & 0.258 & 57.19s & -3.87pp & 3.6$\times$ \\
& & Verification Self-Correction & 0.374 & 79.97s & +7.74pp & 5.0$\times$ \\
\midrule
Qwen2.5-Coder-3B & TBG & Adaptive Decoding & 0.252 & 30.74s & -4.52pp & 1.9$\times$ \\
\midrule
\multirow{3}{*}{DeepSeek-Coder-1.3B}
& \multirow{3}{*}{SLT}
& Baseline (greedy decoding) & 0.181 & 8.29s & -- & 1.0$\times$ \\
& & Uncertainty Self-Correction & 0.239 & 20.53s & +5.81pp & 2.5$\times$ \\
& & Verification Self-Correction & 0.381 & 41.63s & +20.00pp & 5.0$\times$ \\
\midrule
DeepSeek-Coder-1.3B & TBG & Adaptive Decoding & 0.187 & 21.28s & +0.65pp & 2.7$\times$ \\
\bottomrule
\end{tabular}%
}
\caption{Self-correction results on \textbf{BigCodeBench} ($n{=}155$ tasks, v0.1.4 split) across models, feature settings, and decoding strategies. The HumanEval pattern replicates: verification-based self-correction is the only method to improve Pass@1 in all three configurations, while adaptive decoding degrades it in two of three and uncertainty-based self-correction in two of three. Baselines are substantially lower than on HumanEval (0.18--0.30 vs.\ 0.38--0.81), reflecting the higher difficulty of BigCodeBench. Latency conventions follow Table~\ref{tab:humaneval}.}
\label{tab:bigcodebench_results}
\end{table*}

\section*{Limitations and Future Work}

Our study has several limitations. First, all experiments use small models (1.3B–3B parameters); larger models may exhibit different uncertainty profiles and benefit differently from self-correction policies. Second, while we evaluate on both HumanEval and BigCodeBench, real-world code generation involves longer programs, partial specifications, and incomplete tests, where uncertainty estimation and verification are both more challenging and more valuable. Third, proactive regeneration currently only supports adaptive decoding; integrating it with verification-based correction remains unexplored. In future work, we plan to scale experiments to larger models and more diverse benchmarks, explore hybrid strategies that combine proactive uncertainty gating with selective verification-based correction, and investigate uncertainty calibration under distribution shift across programming languages and task complexities.





\nocite{langley00}

\bibliography{example_paper}

@article{fomicheva2020unsupervised,
  title={Unsupervised Quality Estimation for Neural Machine Translation},
  author={Fomicheva, Marina and Sun, Shuo and Yankovskaya, Lisa and Blain, Fr{\'e}d{\'e}ric and Guzm{\'a}n, Francisco and Fishel, Mark and Aletras, Nikolaos and Chaudhary, Vishrav and Specia, Lucia},
  journal={Transactions of the Association for Computational Linguistics},
  volume={8},
  pages={539--555},
  year={2020},
  publisher={MIT Press},
  url={https://arxiv.org/abs/2005.10608},
  doi={10.1162/tacl_a_00330}
}

@inproceedings{tonolini2024bayespe,
    author = {Tonolini, Francesco and Massiah, Jordan and Aletras, Nikolaos and Kazai, Gabriella},
    title = {Bayesian Prompt Ensembles: Model Uncertainty Estimation for Black-Box Large Language Models},
    booktitle = {Findings of the Association for Computational Linguistics: ACL 2024},
    year = {2024}
}

@article{tian2023justask,
  title={Just Ask for Calibration: Strategies for Eliciting Calibrated Confidence Scores from Language Models Fine-Tuned with Human Feedback},
  author={Tian, Katherine and Mitchell, Eric and Munro, Robert and Finn, Chelsea},
  journal={arXiv preprint arXiv:2305.14975},
  year={2023},
  url={https://arxiv.org/abs/2305.14975}
}

@article{lin2022teaching,
  title={Teaching Models to Express Their Uncertainty in Words},
  author={Lin, Stephanie and Hilton, Jacob and Evans, Owain},
  journal={arXiv preprint arXiv:2205.14334},
  year={2022},
  url={https://arxiv.org/abs/2205.14334}
}

@article{xiong2023express,
  title={Can Large Language Models Express Their Uncertainty? An Empirical Evaluation of Confidence Elicitation in LLMs},
  author={Xiong, Wenhan and Liang, Percy and Ermon, Stefano},
  journal={arXiv preprint arXiv:2306.13063},
  year={2023},
  url={https://arxiv.org/abs/2306.13063}
}

@article{kadavath2022language,
  title={Language Models (Mostly) Know What They Know},
  author={Kadavath, Saurav and Conerly, Thomas and Askell, Amanda and Henighan, Tom and Jones, Andy and Joseph, Nicholas and Kaplan, Jared and Markov, Todor and McCandlish, Sam and Perez, Ethan and others},
  journal={arXiv preprint arXiv:2207.05221},
  year={2022},
  url={https://arxiv.org/abs/2207.05221}
}

@article{yang2024verbalized,
  title={On Verbalized Confidence Scores for Large Language Models},
  author={Yang, Daniel and Tsai, Yao-Hung Hubert and Yamada, Makoto},
  journal={arXiv preprint arXiv:2412.14737},
  year={2024},
  url={https://arxiv.org/abs/2412.14737}
}

@misc{kuhn2023semanticuncertaintylinguisticinvariances,
      title={Semantic Uncertainty: Linguistic Invariances for Uncertainty Estimation in Natural Language Generation}, 
      author={Lorenz Kuhn and Yarin Gal and Sebastian Farquhar},
      year={2023},
      eprint={2302.09664},
      archivePrefix={arXiv},
      primaryClass={cs.CL},
      url={https://arxiv.org/abs/2302.09664},
}

@misc{li2025semanticvolumequantifyingdetecting,
      title={Semantic Volume: Quantifying and Detecting both External and Internal Uncertainty in LLMs}, 
      author={Xiaomin Li and Zhou Yu and Ziji Zhang and Yingying Zhuang and Swair Shah and Narayanan Sadagopan and Anurag Beniwal},
      year={2025},
      eprint={2502.21239},
      archivePrefix={arXiv},
      primaryClass={cs.CL},
      url={https://arxiv.org/abs/2502.21239}, 
}

@misc{kossen2024semanticentropyprobesrobust,
      title={Semantic Entropy Probes: Robust and Cheap Hallucination Detection in LLMs}, 
      author={Jannik Kossen and Jiatong Han and Muhammed Razzak and Lisa Schut and Shreshth Malik and Yarin Gal},
      year={2024},
      eprint={2406.15927},
      archivePrefix={arXiv},
      primaryClass={cs.CL},
      url={https://arxiv.org/abs/2406.15927}, 
}

@misc{sharma2025assessingcorrectnessllmbasedcode,
      title={Assessing Correctness in LLM-Based Code Generation via Uncertainty Estimation}, 
      author={Arindam Sharma and Cristina David},
      year={2025},
      eprint={2502.11620},
      archivePrefix={arXiv},
      primaryClass={cs.SE},
      url={https://arxiv.org/abs/2502.11620}, 
}

@misc{zhu2025uncertaintyguidedchainofthoughtcodegeneration,
      title={Uncertainty-Guided Chain-of-Thought for Code Generation with LLMs}, 
      author={Yuqi Zhu and Ge Li and Xue Jiang and Jia Li and Hong Mei and Zhi Jin and Yihong Dong},
      year={2025},
      eprint={2503.15341},
      archivePrefix={arXiv},
      primaryClass={cs.SE},
      url={https://arxiv.org/abs/2503.15341}, 
}

@misc{he2025bettercodegenerationadaptive,
      title={Towards Better Code Generation: Adaptive Decoding with Uncertainty Guidance}, 
      author={Kaifeng He and Mingwei Liu and Chong Wang and Zike Li and Yanlin Wang and Xin Peng and Zibin Zheng},
      year={2025},
      eprint={2506.08980},
      archivePrefix={arXiv},
      primaryClass={cs.SE},
      url={https://arxiv.org/abs/2506.08980}, 
}

@misc{brown_verifiers_2025,
  author       = {William Brown},
  title        = {{Verifiers}: Environments for LLM Reinforcement Learning},
  howpublished = {\url{https://github.com/PrimeIntellect-ai/verifiers}},
  year         = {2025}
}

@inproceedings{Ball2015DeconstructingDS,
  title={Deconstructing Dynamic Symbolic Execution},
  author={Thomas Ball and Jakub Daniel},
  booktitle={Dependable Software Systems Engineering},
  year={2015},
  url={https://api.semanticscholar.org/CorpusID:15763279}
}

@article{vashurin-etal-2025-benchmarking,
    title = "Benchmarking Uncertainty Quantification Methods for Large Language Models with {LM}-Polygraph",
    author = "Vashurin, Roman  and
      Fadeeva, Ekaterina  and
      Vazhentsev, Artem  and
      Rvanova, Lyudmila  and
      Vasilev, Daniil  and
      Tsvigun, Akim  and
      Petrakov, Sergey  and
      Xing, Rui  and
      Sadallah, Abdelrahman  and
      Grishchenkov, Kirill  and
      Panchenko, Alexander  and
      Baldwin, Timothy  and
      Nakov, Preslav  and
      Panov, Maxim  and
      Shelmanov, Artem",
    journal = "Transactions of the Association for Computational Linguistics",
    volume = "13",
    year = "2025",
    address = "Cambridge, MA",
    publisher = "MIT Press",
    url = "https://aclanthology.org/2025.tacl-1.11/",
    doi = "10.1162/tacl_a_00737",
    pages = "220--248"
}

@misc{xia2025surveyuncertaintyestimationmethods,
      title={A Survey of Uncertainty Estimation Methods on Large Language Models}, 
      author={Zhiqiu Xia and Jinxuan Xu and Yuqian Zhang and Hang Liu},
      year={2025},
      eprint={2503.00172},
      archivePrefix={arXiv},
      primaryClass={cs.CL},
      url={https://arxiv.org/abs/2503.00172}, 
}

@misc{tao2025revisitinguncertaintyestimationcalibration,
      title={Revisiting Uncertainty Estimation and Calibration of Large Language Models}, 
      author={Linwei Tao and Yi-Fan Yeh and Minjing Dong and Tao Huang and Philip Torr and Chang Xu},
      year={2025},
      eprint={2505.23854},
      archivePrefix={arXiv},
      primaryClass={cs.CL},
      url={https://arxiv.org/abs/2505.23854}, 
}

@article{Huang_2025,
   title={Look Before You Leap: An Exploratory Study of Uncertainty Analysis for Large Language Models},
   volume={51},
   ISSN={2326-3881},
   url={http://dx.doi.org/10.1109/TSE.2024.3519464},
   DOI={10.1109/tse.2024.3519464},
   number={2},
   journal={IEEE Transactions on Software Engineering},
   publisher={Institute of Electrical and Electronics Engineers (IEEE)},
   author={Huang, Yuheng and Song, Jiayang and Wang, Zhijie and Zhao, Shengming and Chen, Huaming and Juefei-Xu, Felix and Ma, Lei},
   year={2025},
   month=feb, pages={413–429} }

@article{bigcodebench,
  title={BigCodeBench: Benchmarking Code Generation with Diverse Function Calls and Complex Instructions},
  author={Zhuo, Terry Yue and others},
  journal={arXiv preprint arXiv:2406.15079},
  year={2024}
}

@inproceedings{fadeeva-etal-2023-lm,
    title = "{LM}-Polygraph: Uncertainty Estimation for Language Models",
    author = "Fadeeva, Ekaterina  and
      Vashurin, Roman  and
      Tsvigun, Akim  and
      Vazhentsev, Artem  and
      Petrakov, Sergey  and
      Fedyanin, Kirill  and
      Vasilev, Daniil  and
      Goncharova, Elizaveta  and
      Panchenko, Alexander  and
      Panov, Maxim  and
      Baldwin, Timothy  and
      Shelmanov, Artem",
    editor = "Feng, Yansong  and
      Lefever, Els",
    booktitle = "Proceedings of the 2023 Conference on Empirical Methods in Natural Language Processing: System Demonstrations",
    month = dec,
    year = "2023",
    address = "Singapore",
    publisher = "Association for Computational Linguistics",
    url = "https://aclanthology.org/2023.emnlp-demo.41/",
    doi = "10.18653/v1/2023.emnlp-demo.41",
    pages = "446--461"
}
\bibliographystyle{icml2026}

\newpage
\appendix
\onecolumn
\section{Appendix A}
\subsection{Uncertainty Analysis Details}
This appendix presents uncertainty distributions for each model, showing the relationship between Mean Token Entropy (uncertainty) and correctness.

\begin{figure}[H]
\centering
\includegraphics[width=0.9\textwidth]{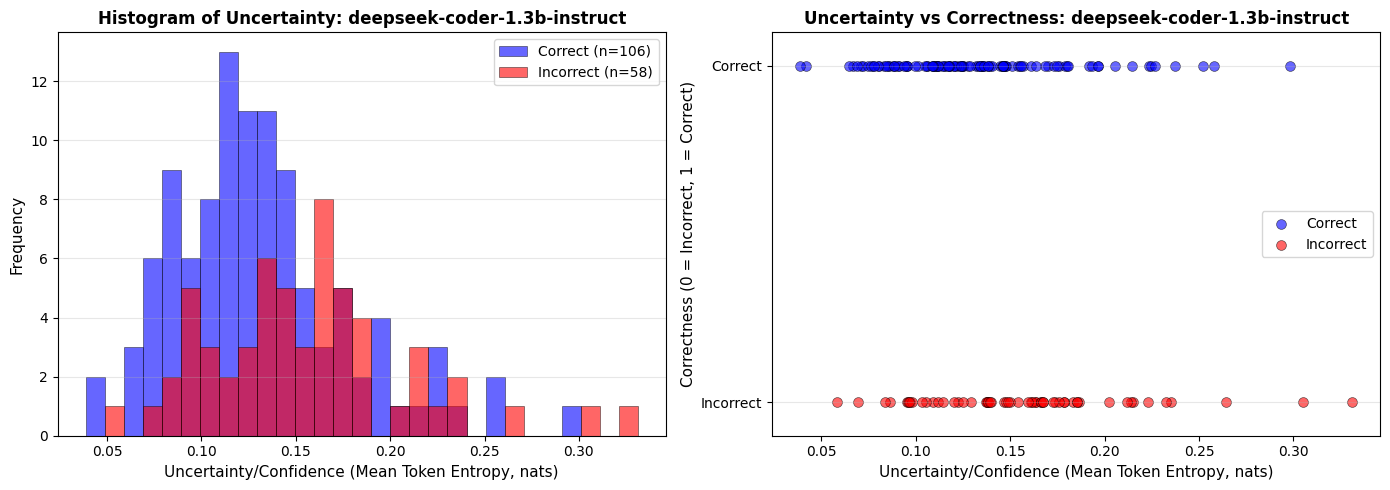}
\caption{DeepSeek Uncertainty Analysis. Correct generations (n=106) peak at low uncertainty (0.10--0.12 nats), while incorrect ones (n=58) show broader distribution with overlap, requiring careful threshold calibration.}
\label{fig:deepseek_uncert}
\end{figure}

\vspace{1cm}

\begin{figure}[H]
\centering
\includegraphics[width=0.9\textwidth]{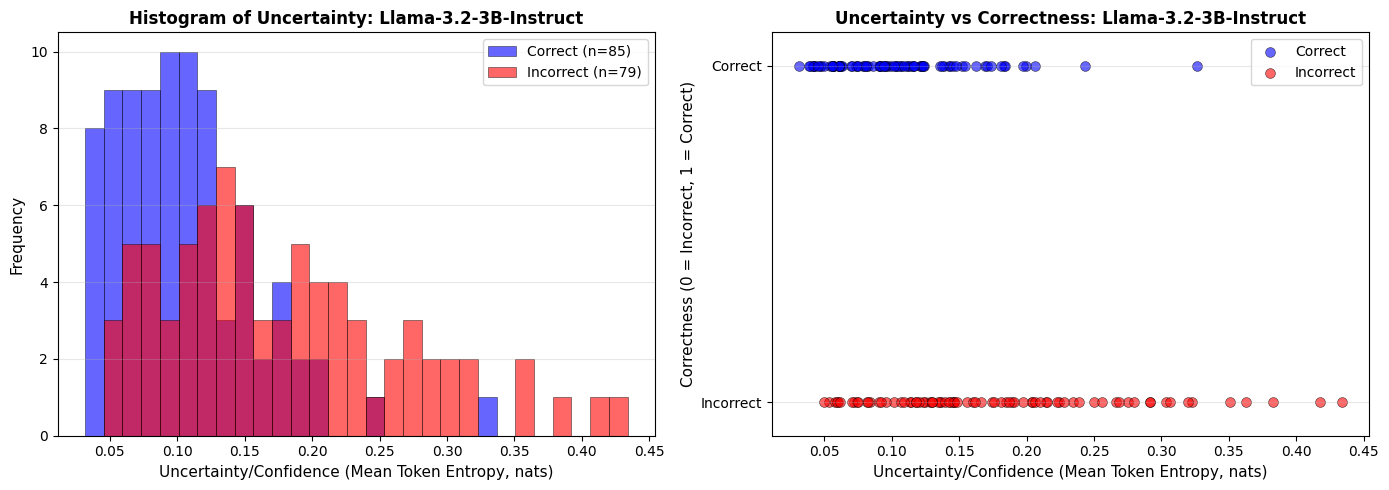}
\caption{Llama Uncertainty Analysis. High overconfidence in incorrect predictions (n=79 vs n=85 correct) with substantial overlap at low uncertainty explains the mixed results with adaptive decoding.}
\label{fig:llama_uncert}
\end{figure}

\vspace{1cm}

\begin{figure}[H]
\centering
\includegraphics[width=0.9\textwidth]{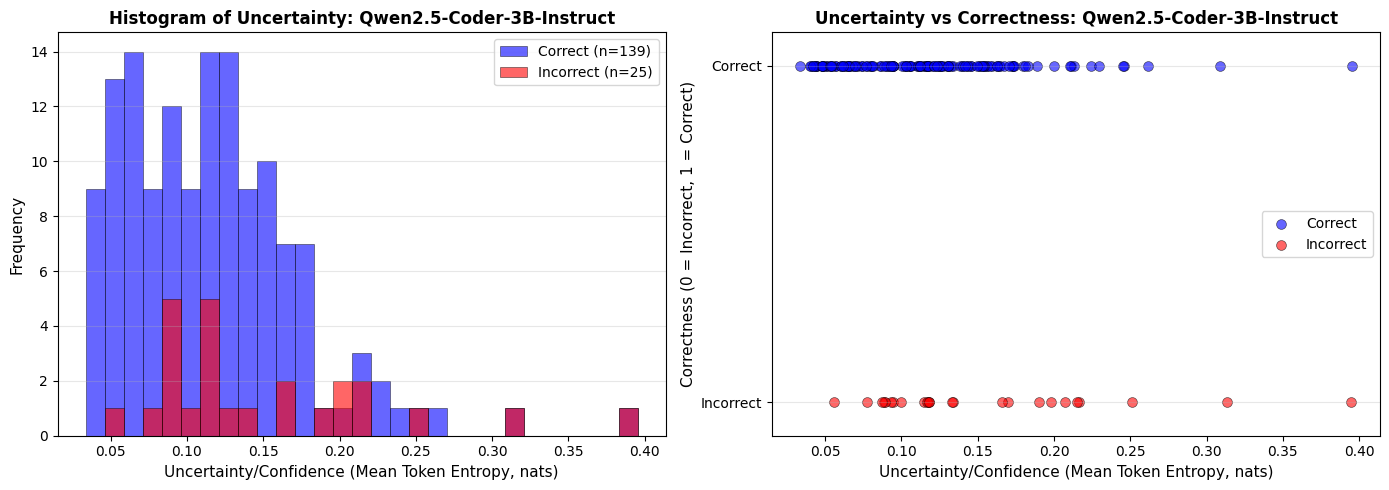}
\caption{Qwen Uncertainty Analysis. Well-calibrated uncertainty profile with correct generations (n=139) concentrated at low uncertainty and few incorrect samples (n=25) supports effective self-correction strategies.}
\label{fig:qwen_uncert}
\end{figure}

\begin{figure}[H]
    \centering
    \includegraphics[width=\linewidth]{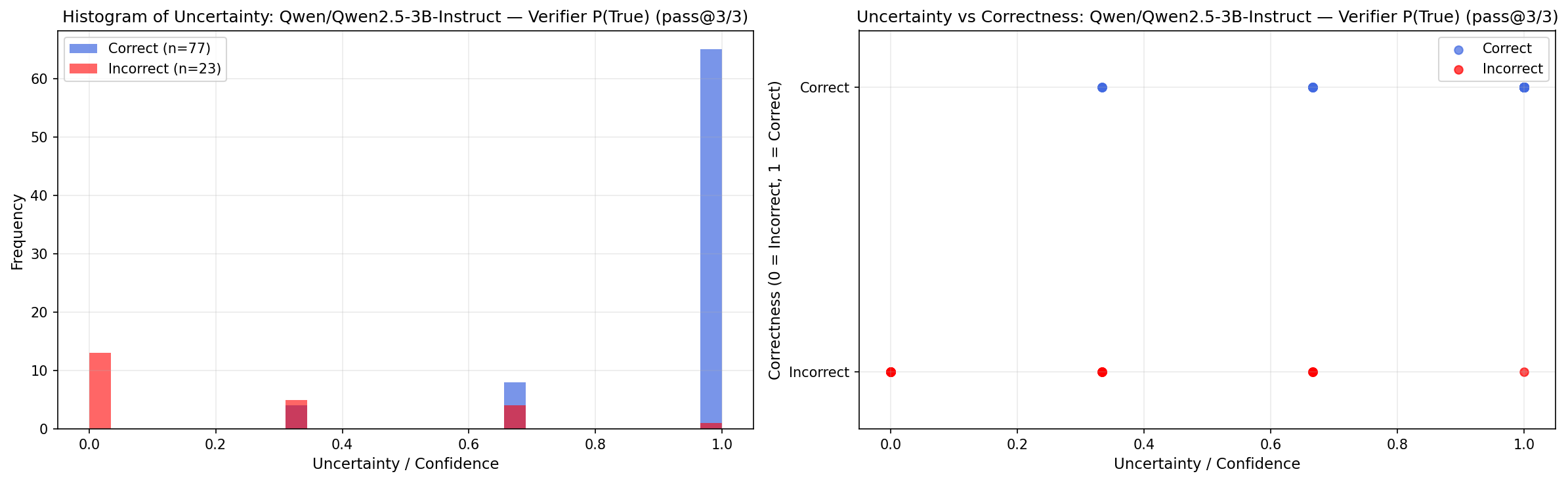} 
    \caption{
    Qwen2.5-3B-Instruct P(True) uncertainty analysis using pass@3 verification.
    Correct generations ($n=77$) are concentrated at high confidence values, while most incorrect samples ($n=23$) receive low P(True), indicating well calibrated uncertainty estimates.
    }
    \label{fig:qwen_ptrue}
\end{figure}

\begin{figure}[H]
    \centering
    \includegraphics[width=\linewidth]{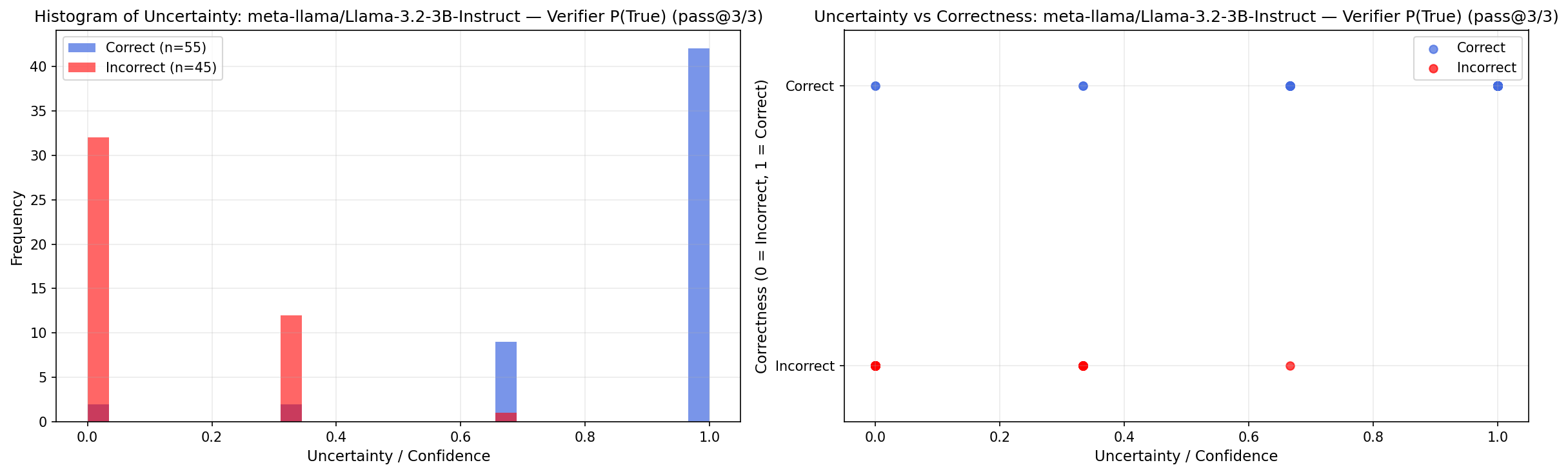}
    \caption{
    Llama-3.2-3B-Instruct P(True) uncertainty analysis using pass@3 verification.
    Correct generations ($n=55$) receive consistently higher confidence than incorrect ones ($n=45$), demonstrating effective uncertainty separation.
    }
    \label{fig:llama_ptrue}
\end{figure}

\begin{figure}[H]
    \centering
    \includegraphics[width=\linewidth]{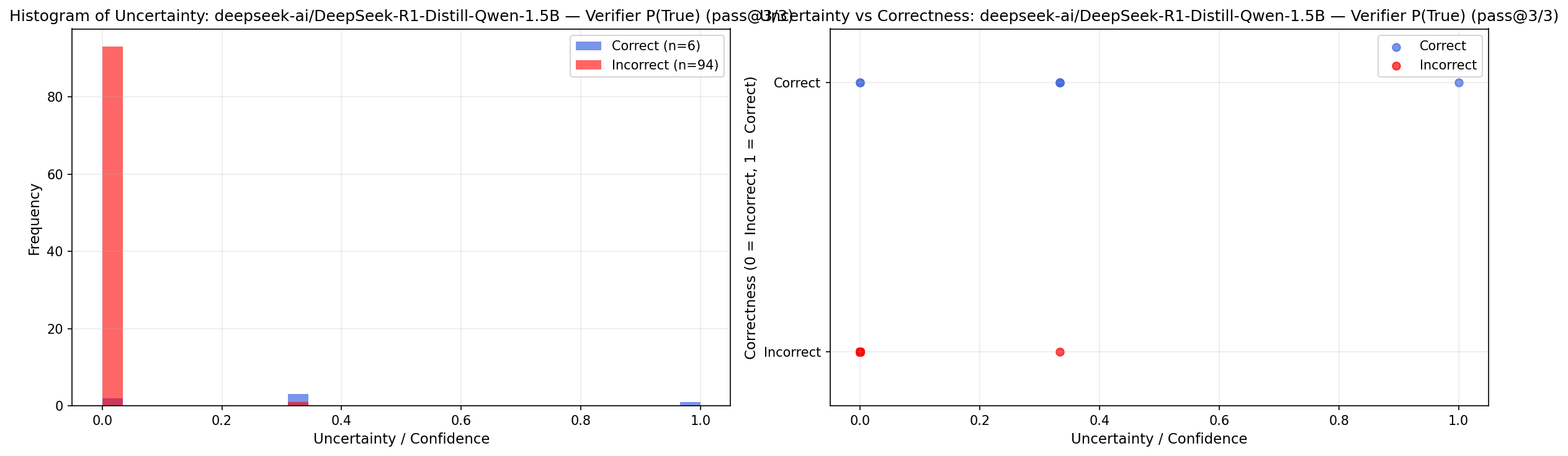}
    \caption{
    DeepSeek-R1-Distill-Qwen-1.5B P(True) uncertainty analysis using pass@3 verification.
    Despite low overall accuracy ($n=6$ correct, $n=94$ incorrect), the verifier assigns near zero confidence to most incorrect solutions, demonstrating informative uncertainty estimates.
    }
    \label{fig:deepseek_ptrue}
\end{figure}

\begin{figure}[H]
\centering
\includegraphics[width=0.9\textwidth]{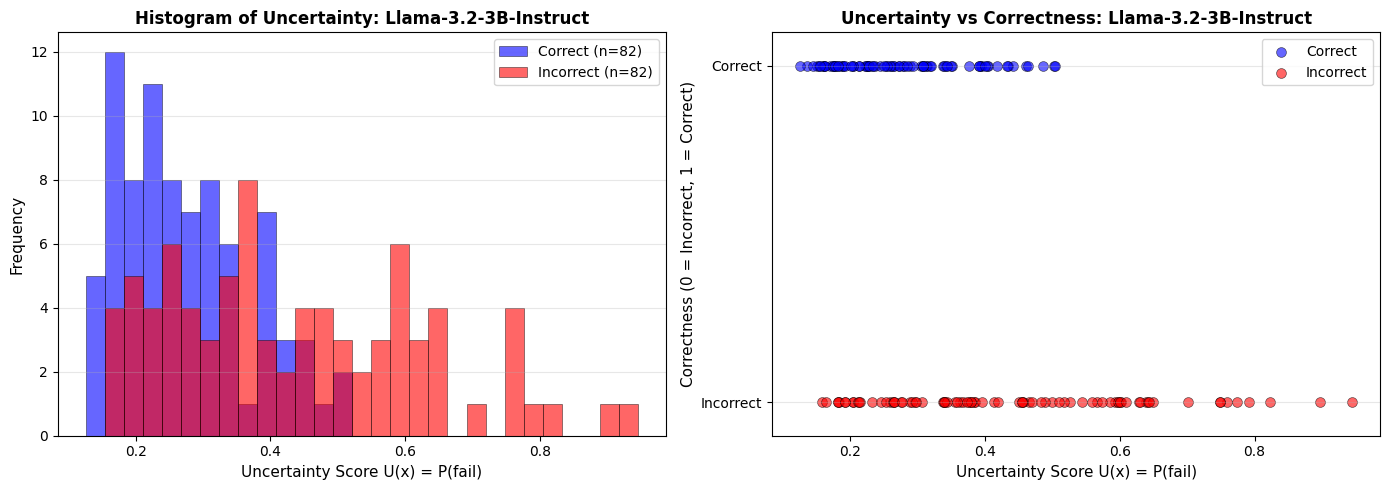}
\caption{Llama-3.2-3B-Instruct uncertainty analysis. Correct and incorrect generations show substantial overlap at low predicted failure probability, indicating overconfident uncertainty estimates and limited separability for threshold-based self-correction.}
\label{fig:llama_bayes_uncert}
\end{figure}

\begin{figure}[H]
\centering
\includegraphics[width=0.9\textwidth]{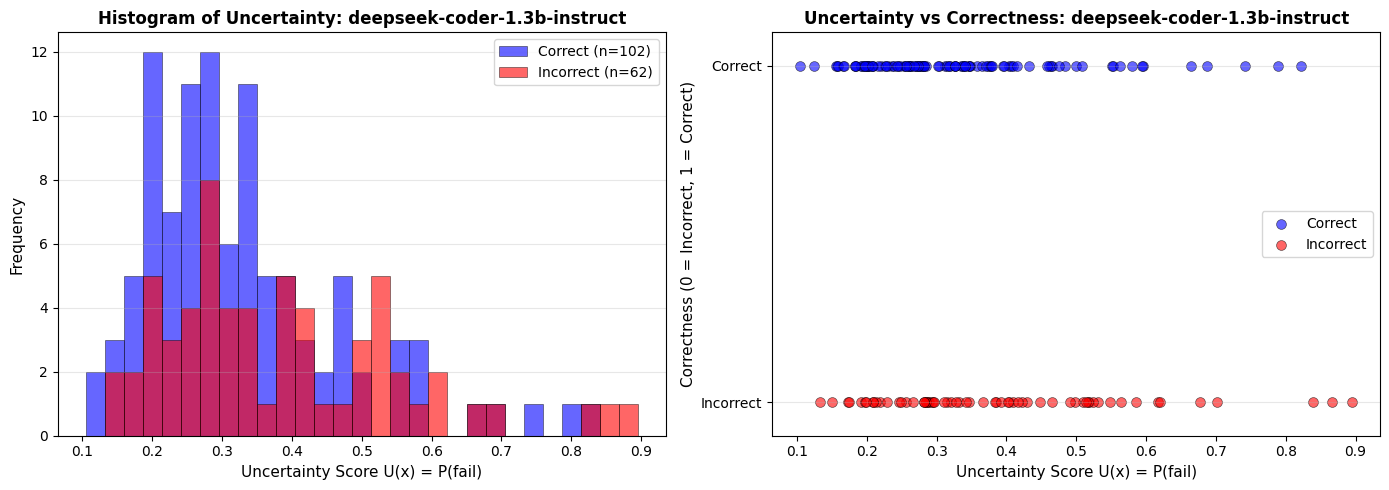}
\caption{DeepSeek-Coder-1.3B-Instruct uncertainty analysis. Correct generations concentrate at lower uncertainty, while incorrect outputs exhibit a broader distribution with overlap, requiring careful uncertainty threshold calibration.}
\label{fig:deepseek_bayes_uncert}
\end{figure}

\subsection{Prompt Templates}
\label{prompts}
This appendix contains the prompt templates used for uncertainty
estimation methods where prompts significantly influence model behavior.
\subsubsection{Mean Token Entropy (MTE)}
\paragraph{Code Generation Prompt (MTE).}

MTE uses the same code generation prompts as baseline greedy decoding, with model-specific formatting:

\begin{itemize}
    \item \textbf{Llama and DeepSeek models:} Use chat format with system instruction: "You are a strict coding assistant. Output only valid Python code for the function, no explanations." The user prompt is the HumanEval problem statement followed by "\textbackslash n\textbackslash n\# Your code below:\textbackslash n".
    
    \item \textbf{Qwen models:} Use plain text format with the HumanEval problem statement followed by "\textbackslash n\# Your code below:\textbackslash n".
\end{itemize}

\subsubsection{Verbalized Confidence Prompts}
\label{app:verbalized-confidence-prompts}

This section documents the exact prompts used for verbalized confidence based uncertainty estimation, where models explicitly report a scalar confidence score for each generated candidate program.

\paragraph{System Prompt (Code Generation).}
All model generations, including greedy decoding, were conditioned on the following system instruction:

\begin{quote}
You are a Python coding assistant. Complete the function so that it passes the tests. Return only Python code, no explanation.
\end{quote}

\paragraph{User Prompt (HumanEval Task Prompt).}
For each task, the user prompt was the HumanEval problem statement and function signature provided by the benchmark, denoted as \texttt{row["prompt"]}. Candidate solutions were evaluated by executing the concatenation of the HumanEval prompt, the model generated code, and the benchmark unit tests, denoted as \texttt{row["test"]}, under a timeout constraint.

\paragraph{Self Evaluation Prompt (Verbalized Confidence).}
To obtain verbalized confidence scores, we prompt the model to assess its confidence that a given candidate solution will pass all tests. The self evaluation prompt is:

\begin{quote}
Problem:\\
\texttt{[HumanEval problem prompt]}

Candidate solution:\\
\texttt{```python}\\
\texttt{[model generated code]}\\
\texttt{```}

Question: How confident are you (0--100) that this solution is correct and will pass all test cases?\\
Respond with only a number between 0 and 100:
\end{quote}

Additionally, during decoding for this self evaluation step, the system message was set to:

\begin{quote}
You are an expert code evaluator. You must assess your confidence that the given code solution is correct and will pass all test cases. Respond with ONLY a number between 0 and 100, where 0 means absolutely certain the code is wrong and 100 means absolutely certain the code is correct.
\end{quote}

\subsubsection{P(True) Prompts}

\label{app:ptrue-prompts}

This section documents the exact prompts used to generate candidate programs, elicit self evaluated confidence, and optionally trigger regeneration for uncertainty guided correction.

\paragraph{System Prompt (Code Generation).}
All model generations, including greedy decoding and stochastic rollouts, were conditioned on the following system instruction:

\begin{quote}
You are a Python coding assistant. Complete the function so that it passes the tests. Return only Python code, no explanation.
\end{quote}

\paragraph{User Prompt (HumanEval Task Prompt).}
For each task, the user prompt was the HumanEval problem statement and function signature provided by the benchmark, denoted as \texttt{row["prompt"]} in the implementation. Candidate solutions were evaluated by executing the concatenation of the HumanEval prompt, the model generated code, and the benchmark unit tests, denoted as \texttt{row["test"]}, under a timeout constraint. See the released code for implementation details.

\paragraph{Self Evaluation Prompt (Baseline P(True)).}
To obtain self evaluated correctness, we prompt the model to judge whether a given candidate solution will pass all tests. The self evaluation prompt is:

\begin{quote}
You are a strict evaluator for Python coding tasks.\\
Answer with exactly one word: True or False.

Problem:\\
\texttt{[HumanEval problem prompt]}

Candidate solution:\\
\texttt{```python}\\
\texttt{[model generated code]}\\
\texttt{```}

Question: Does the candidate solution pass all tests?\\
Answer:
\end{quote}

Additionally, during decoding for this self evaluation step, the system message was set to:

\begin{quote}
Answer strictly True or False.
\end{quote}

\paragraph{Regeneration Prompt (Optional Self Correction).}
For uncertainty guided regeneration, when enabled, we append the following instruction to the original HumanEval prompt:

\begin{quote}
The previous attempt failed. Try a different approach.
\end{quote}

The regenerated program is then evaluated using the same external verifier, and a new verifier based P(True) is estimated via stochastic rollouts.

\section{APPENDIX B}

\subsection{Probe Design Choices}

\label{app:probe-design}

Our semantic entropy probes (SEP) employ several design choices that improve uncertainty estimation accuracy:

\textbf{Multi-layer feature concatenation:} We extract and concatenate features from the last three transformer layers ($[-3, -2, -1]$), creating 6144--9216 dimensional vectors that capture both syntactic (earlier layers) and semantic (later layers) information. This multi-layer approach improves probe accuracy compared to single-layer features.


\textbf{MLP classifier:} We use a 3-layer MLP ($256 \rightarrow 128 \rightarrow 64$) rather than linear models, enabling non-linear feature interactions that improve accuracy (test AUROC: 0.62--0.89). Early stopping prevents overfitting.

\textbf{Feature standardization:} Features are standardized to ensure equal contribution from all layers.

\textbf{Model-specific thresholds:} F1-optimized thresholds (0.30--0.70) account for varying uncertainty distributions across models, improving precision-recall trade-offs.


\begin{table}[h]
\centering
\caption{Semantic Probe Test Set Performance}
\label{tab:probe_accuracy}
\resizebox{0.65\textwidth}{!}{%
\begin{tabular}{lccccc}
\toprule
\textbf{Model} & \textbf{Feature} & \textbf{Acc.} & \textbf{AUROC} & \textbf{F1} & \textbf{Threshold} \\
\midrule
Llama-3.2-3B-Instruct & SLT & 0.60 & 0.75 & 0.69 & 0.5 \\
 & TBG & 0.68 & 0.89 & 0.79 & 0.7 \\
Qwen2.5-Coder-3B-Instruct & SLT & 0.68 & 0.76 & 0.96 & 0.7 \\
 & TBG & 0.56 & 0.67 & 0.85 & 0.3 \\
DeepSeek-Coder-1.3B & SLT & 0.76 & 0.88 & 0.92 & 0.5 \\
 & TBG & 0.64 & 0.62 & 0.78 & 0.3 \\
\bottomrule
\end{tabular}%
}
\end{table}

\subsection{Threshold Tuning}

Threshold selection is critical for uncertainty-guided decoding, as it determines when to trigger adaptive decoding or self-correction. We employ a systematic threshold tuning procedure that evaluates candidate thresholds (0.3, 0.4, 0.5, 0.6, 0.7) on held-out test data, computing precision, recall, F1 score, accuracy, and trigger rate (the percentage of examples that would trigger corrections) at each threshold. The optimal threshold is selected to maximize F1 score, which balances precision (avoiding unnecessary corrections) and recall (capturing truly uncertain generations). This F1-optimized approach yields model- and feature-method-specific thresholds ranging from 0.3 to 0.7, reflecting the varying uncertainty distributions across models. For instance, Qwen2.5-Coder-3B-Instruct with SLT features uses a threshold of 0.7 (F1: 0.96, accuracy: 0.96), while DeepSeek-Coder-1.3B-Instruct with TBG features uses 0.3 (F1: 0.78, accuracy: 0.80). These tuned thresholds significantly outperform a naive median-based approach (which uses the 50th percentile of probe predictions), as the F1-optimized thresholds account for the asymmetric costs of false positives (unnecessary corrections) versus false negatives (missed corrections). The resulting thresholds are automatically loaded during inference, ensuring consistent and optimized uncertainty-guided decisions across all experiments.

\end{document}